\documentclass[lettersize,journal]{IEEEtran}
\usepackage{amsmath,amsfonts}
\usepackage{algorithmic}
\usepackage{algorithm}
\usepackage{array}
\usepackage[caption=false,font=normalsize,labelfont=sf,textfont=sf]{subfig}
\usepackage{textcomp}
\usepackage{stfloats}
\usepackage{url}
\usepackage{verbatim}
\usepackage{graphicx}
\usepackage{cite}
\usepackage{multirow}
\usepackage{makecell}
\usepackage{float}
\begin{document}
\bibliographystyle{IEEEtran}

% 1）对标题进行了修改
\title{FALSE: False Negative Samples Aware Contrastive Learning for Semantic Segmentation of High-Resolution Remote Sensing Image}

\author{
    Zhaoyang Zhang,
    Xuying Wang, 
    Xiaoming Mei,
    Chao Tao,
    Haifeng~Li*
    % <-this % stops a space
\thanks{This work was supported by the National Natural Science Foundation of China(41871364, 41861048)and by the High Performance Computing Platform of Central South University. Corresponding author: lihaifeng@csu.edu.cn.}% <-this % stops a space
\thanks{
Z. Zhang, X. Wang, X. Mei, C. Tao, H. Li is with the School of Geosciences and Info-Physics, Central South University.}
}

% The letter headers
\markboth{Journal of \LaTeX\ Class Files,~Vol.~14, No.~8, August~2021}%
{Shell \MakeLowercase{\textit{et al.}}: A Sample Article Using IEEEtran.cls for IEEE Journals}

\maketitle

\begin{abstract}
Self-supervised contrastive learning (SSCL) is a potential learning paradigm for learning remote sensing image (RSI)-invariant features through the label-free method. The existing SSCL of RSI is built based on constructing positive and negative sample pairs. However, due to the richness of RSI ground objects and the complexity of the RSI contextual semantics, the same RSI patches have the coexistence and imbalance of positive and negative samples, which causing the SSCL pushing negative samples far away while pushing positive samples far away, and vice versa. We call this the sample confounding issue (SCI). To solve this problem, we propose a False negAtive sampLes aware contraStive lEarning model (FALSE) for the semantic segmentation of high-resolution RSIs. Since the SSCL pretraining is unsupervised, the lack of definable criteria for false negative sample (FNS) leads to theoretical undecidability, we designed two steps to implement the FNS approximation determination: coarse determination of FNS and precise calibration of FNS. We achieve coarse determination of FNS by the FNS self-determination (FNSD) strategy and achieve calibration of FNS by the FNS confidence calibration (FNCC) loss function. Experimental results on three RSI semantic segmentation datasets demonstrated that the FALSE effectively improves the accuracy of the downstream RSI semantic segmentation task compared with the current three models, which represent three different types of SSCL models. The mean Intersection-over-Union on ISPRS Potsdam dataset is improved by 0.7\% on average; on CVPR DGLC dataset is improved by 12.28\% on average; and on Xiangtan dataset this is improved by 1.17\% on average. This indicates that the SSCL model has the ability to self-differentiate FNS and that the FALSE effectively mitigates the SCI in self-supervised contrastive learning.
\end{abstract}


\IEEEpeerreviewmaketitle

\vspace{-5mm}
\section{Introduction}

\begin{figure}
\centering
\includegraphics[width=8.7cm]{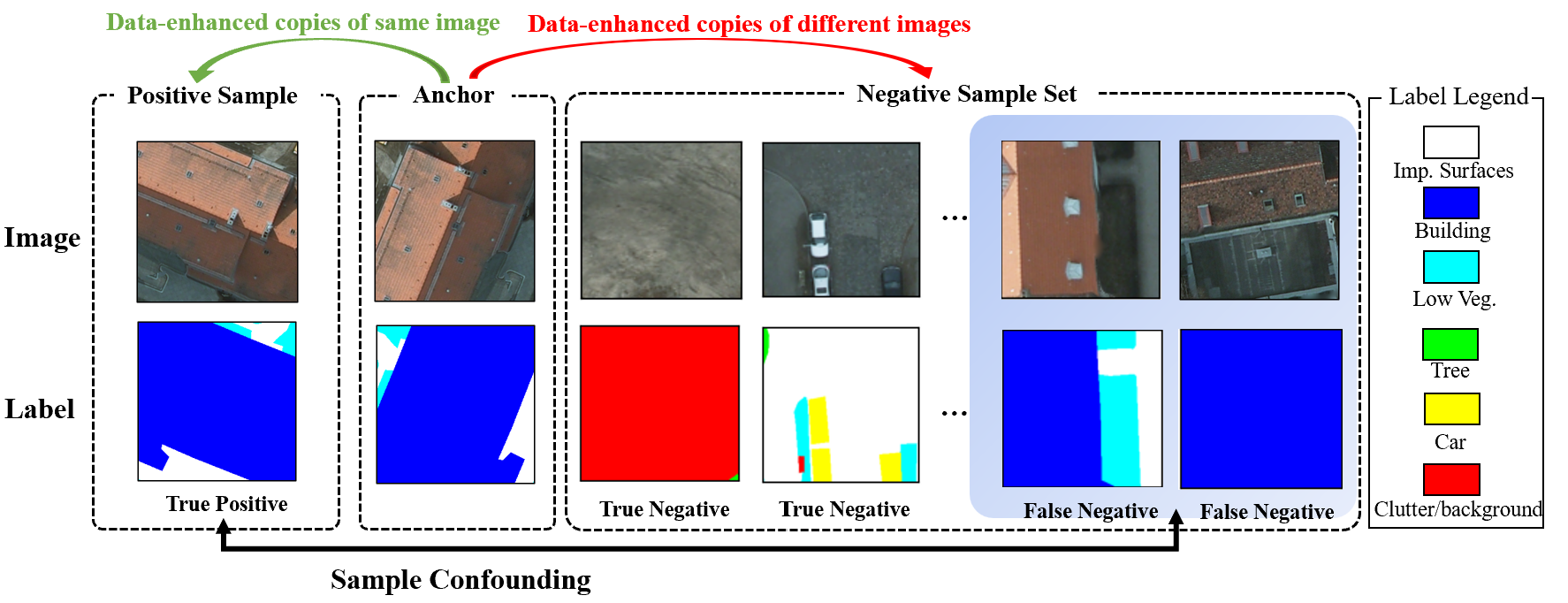}
\vspace{-3mm}
\caption{Example of false negative sample (FNS) and sample confounding issue (SCI) in SSCL for semantic segmentation of high-resolution RSI. SCI will arise when model pushes negative sample image patch that contains positive samples far away.}
\label{fig_sci}
\end{figure}
%\vspace{-8mm}

\IEEEPARstart{D}{eep} neural networks (DNNs) trained in a supervised learning manner have made remarkable progress in remote sensing image (RSI) scene classification\cite{sup_class}, target detection\cite{sup_det}, and semantic segmentation\cite{sup_seg0,sup_seg}. The dependence of this approach on massive, high-quality labeled samples has become a bottleneck for wide-scale application\cite{m_contrastive,sup_cha1,sup_cha2}. The promise of self-supervised contrastive learning (SSCL) has made it possible to learn the RSI invariant features from massive unlabeled data\cite{m2_contrastive,rs_ssl_cl,rs_ssl_seg_gl,simclr}.

The core idea of SSCL is to cleverly obtain copies of the same image patches as positive samples and other different images as negative samples by data augmentation methods of spatial and spectral transformations such as rotation, scaling, random color distortion, and Gaussian blur \cite{data_aug}, and to construct self-supervised signals by pulling the positive samples closer while pushing the negative samples farther away. This is done instead of manually labeling them as supervised signals, thus forcing DNNs to obtain spatial and spectral invariant representations\cite{moco,simclr,contralea_with_simclr,contralea_with_simclr2}.

However, due to scene complexity, ground object richness, and unbalanced distribution of samples of RSI, there is a phenomenon in which positive and negative samples coexist in the same patch and are highly unbalanced. As shown in Fig 1, the false negative samples for the selected anchor sample in the last two columns contain the same ground features as the positive sample, causing the SSCL pushing negative samples far away while pushing positive samples far away, and vice versa. We call this the sample confounding issue (SCI). The performance loss of the model due to SCI is called the sample confounding effect (SCE), while the negative sample image patch containing positive samples is called the false negative sample (FNS) because it gives the wrong feedback signal to the model.

Current methods to solve the SCE problem in SSCL are mainly considered from the perspective of samples and can be divided into two categories. One category aims to improve the quality of negative sample construction by attaching other unsupervised methods to the original SSCL and using the additional unsupervised results to guide the self-supervised model to construct higher-quality positive and negative sample pairs \cite{pcl,deepcluster,swav}. However, it is often difficult to improve the construction quality of positive and negative sample pairs in RSI processing by using additional unsupervised clustering methods. Nevertheless, this may introduce defects of related unsupervised methods because the ground objects in RSI often have problems such as sample imbalance, intraclass differences, and interclass similarity, leading to the ineffectiveness of unsupervised clustering methods\cite{imbalanced}.
The second category considers abandoning the construction of negative samples \cite{barlowtwins,byol,no_negative_method}, which means that the model's performance will depend only on the construction of positive samples. Considering that the FNS is essentially positive samples in the dataset, this type of approach completely avoids the generation of FNS. Nevertheless, it also means that the model will not use the positive samples that already exist in the dataset, which may reduce the model's ability to extract RSI invariant features.

Unlike the above methods, our observation is as follows: the SSCL of the RSI model itself can distinguish between true and false negative samples. This ability comes from the correct self-supervised signals given to the model by the true positive and true negative samples. This ability is potentially reinforced continuously as the model is trained. We refer to this as the ability of FNS self-determination (FSD). This observation motivates us to rethink the SCE problem from the perspective of the model rather than directly from the perspective of the sample.

The fundamental difficulty of using FSD to determine the FNS is that self-supervised pretraining is essentially an unsupervised process. The lack of definable criteria for the FNS leads to theoretical undecidability, so we can only approximately determine the FNS by some strategy. Approximate determination of FNS can be divided into two steps in terms of process: coarse determination of FNS and precise calibration of FNS. The former is the initial screening of FNS to ensure completeness, and the latter is the precise selection based on the former to ensure accuracy.

We propose the False negAtive sampLe aware contraStive lEarning model (FALSE), which achieves the coarse determination of FNS through the FNS self-determination (FNSD) strategy and achieves the precise calibration of FNS by designing the FNS confidence calibration (FNCC) loss function. In the FNSD strategy, the anchor sample in the closer positive sample pair in the embedding space is used as the benchmark, and the negative sample with the highest similarity to the anchor sample is determined as the possible FNS. The FNCC loss function is designed to improve the contribution of the possible FNS to the positive sample term of the original contrastive loss function\cite{infoNCE} and reduce its contribution to the negative sample term of the loss function to mitigate SCE in the SSCL model. The contributions in this letter are as follows:
\begin{enumerate}
\item{We propose a False negAtive sampLe aware contraStive lEarning model (FALSE) for the semantic segmentation of high-resolution RSIs. FALSE determines the approximate determination of FNS in SSCL from the perspective of the model rather than samples and mitigates the SCI in the SSCL of RSIs.}
\item{We designed the FNS confidence calibration (FNCC) loss function quantitatively rather than qualitatively to characterize the strength of the ability of FNS self-determination (FSD) in the form of confidence weights.}
\item{The experimental results on three semantic segmentation datasets show that FALSE relative to SimCLR, PCL, and Barlow twins improves mean Intersection-over-Union (mIoU) on ISPRS Potsdam dataset by 0.7\% on average on ISPRS Potsdam dataset, improves mIoU by 12.28\% on average on CVPR DGLC dataset, and improves mIoU by 1.17\% on average on Xiangtan dataset.}
\end{enumerate}

\vspace{-5 mm}
\section{Methodology}

\begin{figure*}[ht]
\centering
\includegraphics[width=14cm]{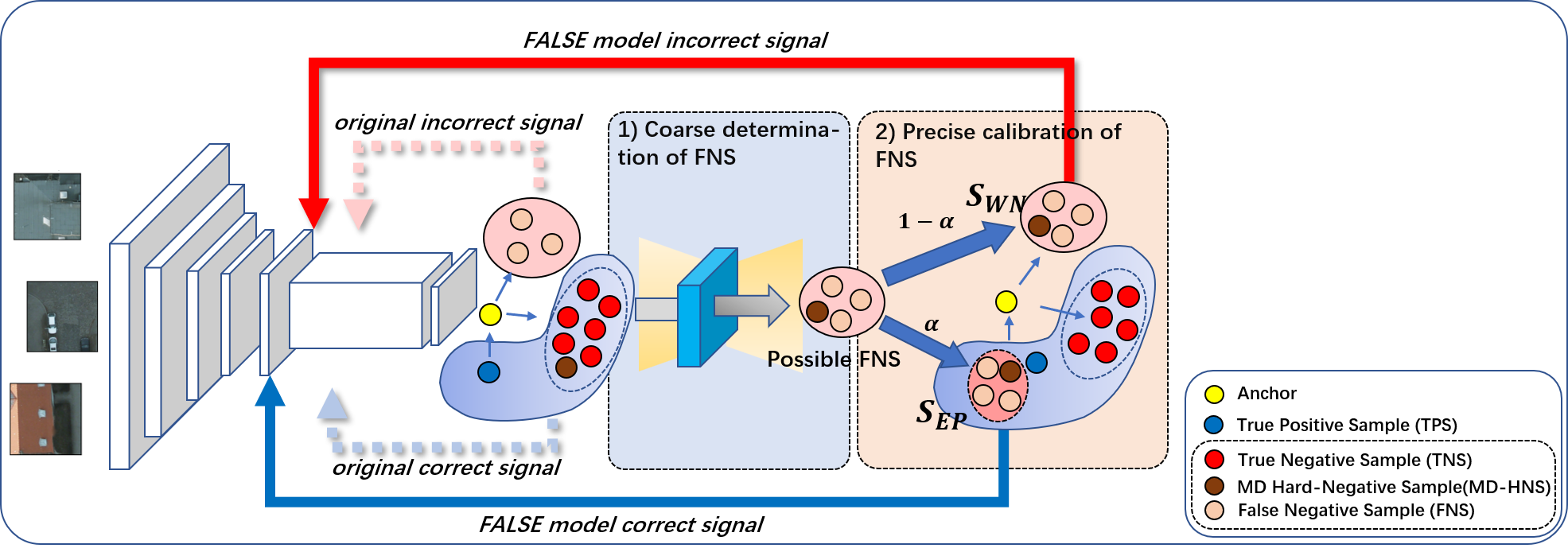}
\vspace{-3mm}
\caption{Overview of FALSE model. Blue square represents FALSE model's FSD module in coarse dermination of FNS}
\label{fig_false}
\end{figure*}

In the SSCL of RSI, the presence of true positive sample (TPS, blue dots in Fig \ref{fig_false}) and true negative sample (TNS, red dots in Fig \ref{fig_false}) will give the model a correct self-supervised signal about the RSI invariant features. In contrast, false negative sample (FNS, pink dots in Fig. \ref{fig_false}) in the negative sample set will give the model an incorrect signal about the RSI invariant features, creating the SCI in SSCL.

Since SSCL pretraining is essentially an unsupervised process, the lack of definable criteria for FNS leads to theoretical undecidability, so we can only approximately determine the FNS by some strategies. The approximate determination of FNS can be divided into two steps: 1) coarse determination of FNS and 2) precise calibration of FNS. The former is the initial screening of FNS to ensure the completeness of FNS; the latter is precise on the former to ensure the accuracy of FNS selection.

\vspace{-3 mm}
\subsection{Coarse determination of FNS}
\subsubsection{Determination benchmark}
Since the goal of the SSCL model is to bring positive samples closer and push negative samples farther, if the model projects a positive sample pair to a closer location in the embedding space, then the model is currently better at learning invariant features about that positive sample pair. Based on this, the anchor sample of the closer positive sample pair in the embedding space is selected as the benchmark for determination, maximizing the use of the feature extraction information that the model has learned, thus minimizing the model's misjudgment.

Suppose the benchmark anchor sample is denoted as $o_{key}$, its corresponding positive sample is denoted as $p$, and $sim(\cdot,\cdot)$ denotes the calculation of the feature similarity between the two samples. Then, $o_{key}$ satisfies the condition that
\begin{equation}
sim\left(o_{key},\ p\right)>T \label{1}
\end{equation}
In Eq. \eqref{1}, $T$ denotes the positive sample pair similarity threshold, which controls the proximity of positive sample pairs in the embedding space.

\subsubsection{Determination condition}
Based on the determination benchmark satisfying Eq. \ref{1}, we calculate the similarity between all negative samples and the benchmark anchor sample $o_{key}$ in the embedding space and determine the negative sample with the highest similarity to the benchmark anchor sample $o_{key}$ as the possible FNS. Suppose $n$ is used to denote the negative samples to be judged, and $n_{pf}$ denotes the possible FNS, the above determination condition can be simply described as:

\begin{equation}
\begin{aligned}
\left|sim(o_{key},n_{pf})-sim(o_{key},p)\right|
\\\rightarrow min\left|sim(o_{key},n)-sim(o_{key},p)\right | \label{2}
\end{aligned}
\end{equation}

\subsubsection{Possible FNS analysis}
Influenced by the SCI, the possible FNS obtained by the FSD is not all FNS but also contains TNS. Nevertheless, since $o_{key}$ represents the best level of the current model's ability to extract image features, it is difficult to use the model's FSD ability to eliminate this part of the TNS from the set of possible FNS. Noting that the indistinguishability of such negative samples is for the model, we refer to this part of the TNS among the possible FNS $n_{pf}$ as the model-dependent hard negative sample (MD-HNS, brown dots in Fig \ref{fig_false}) in the SSCL. Suppose $n_f$ is used to denote the FNS and $n_h$ to denote the MD-HNS, the composition of the possible FNS obtained can be simply described as:
\begin{equation}
n_{pf}=n_f+n_h \label{3}
\end{equation}

\vspace{-3mm}
\subsection{Precise calibration of FNS}
\subsubsection{FNS confidence calibration (FNCC) loss function}
To calibrate the obtained possible FNS and mitigate the impact of MD-HNS on the model performance, we design the FNCC loss function by introducing confidence weights $\alpha$ to calibrate the possible FNS ($n_{pf}$) as positive samples. 

The original loss function of the SSCL of the RSI model mainly consist of two parts\cite{infoNCE}: positive sample term $e^{sim\left(o,p\right)}$ and negative sample term $\sum_{i}^{N}e^{sim\left(o,n^i\right)}$.

And the FNCC multiply the similarity of possible FNS by the confidence weight $\alpha$ to get $S_{EP}$ (see Eq. \eqref{4}), and add it to the original positive sample term of SSCL loss, increases the influence of $n_{pf}$ on the positive sample term of the loss to enhance the correct signal, multiply the similarity of possible FNS by $1-\alpha$ to get $S_{WN}$ (see Eq. \eqref{5}), and replace the similarity corresponding to the possible FNS in the original negative sample term with $S_{WN}$, reduces the influence of $n_{pf}$ on the negative sample term of the loss to weaken the incorrect signal.

When the number of negative samples corresponding to an anchor sample is $N$, the number of possible FNS determined to be obtained is $N_{pf} (N_{pf}<N)$, the FNCC loss function is defined by Eq. \eqref{6}.
\begin{equation}
S_{EP}=\alpha\sum_{j}^{N_{pf}}e^{sim\left(o,n_{pf}^j\right)} \label{4}
\end{equation}

\begin{equation}
S_{WN}=(1-\alpha)\sum_{j}^{N_{pf}}e^{sim\left(o,n_{pf}^j\right)} \label{5}
\end{equation}

\begin{small}
\begin{equation}
\begin{aligned}
L_{FNCC}=-log\frac{e^{sim\left(o,p\right)}+S_{EP}}{e^{sim\left(o,p\right)}+S_{EP}+\sum_{i}^{N-N_{pf}}e^{sim\left(o,n^i\right)}+S_{WN}} \label{6}
\end{aligned}
\end{equation}
\end{small}

In particular, when there is no possible FNS, $N_{pf}=0$, and Eq. \eqref{5} degenerates to original loss function and becomes the original SSCL model.
\subsubsection{Meaning of confidence weights}
The confidence weight $\alpha$ represents the degree of confidence in the model's FSD ability. When $\alpha=0$, the positive sample signal enhancement term and the FNS signal weakening term of the FNCC loss are both 0. The model is the original SSCL model, and the FNSD strategy is not used. When $\alpha=1$, it means that FALSE fully trusts the possible FNS obtained from the model FSD and adjusts all possible FNS to the positive sample, eliminating the contribution of these possible FNS to the negative sample term of the FNCC. When $\alpha$ takes any value less than 1 and greater than 0, it means that the model increases the contribution of possible FNS to the positive sample term of the FNCC from 0 to $\alpha$ times the original contribution to the negative sample term, and weakens its contribution to the negative sample term to $1-\alpha$ times the original contribution.

\vspace{-5mm}
\section{Experimental}

\begin{figure*}[ht]
\centering
\includegraphics[width=14cm]{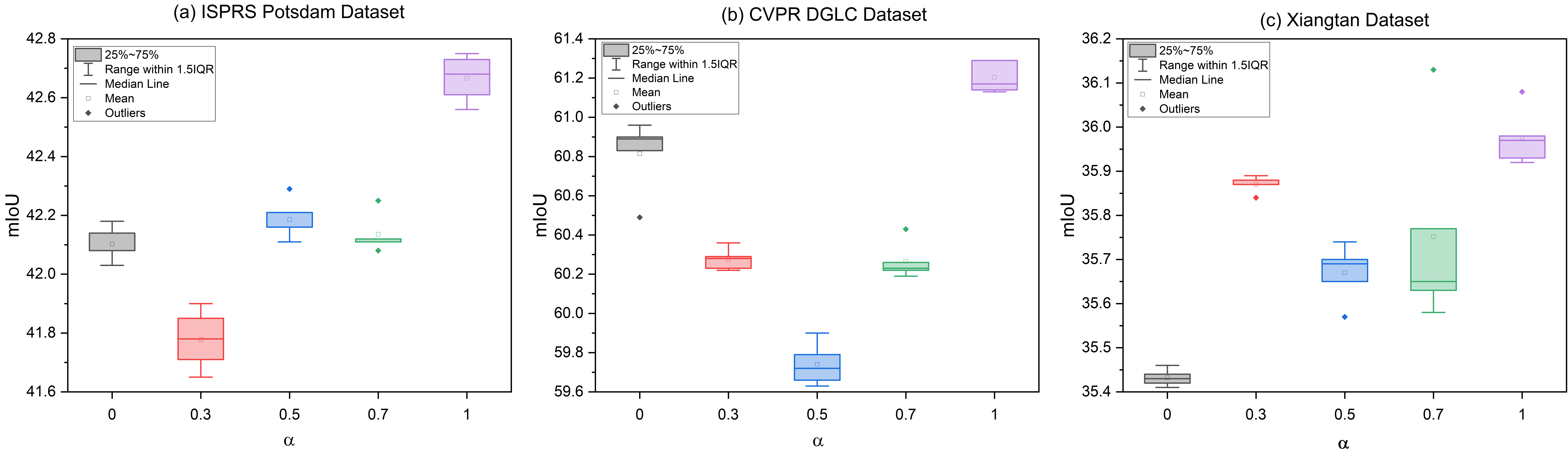}
\vspace{-3mm}
\caption{mIoU (\%) of 3 datasets when five different $\alpha$ are selected. Each confidence weight experiment was repeated five times.}
\label{fig_o}
\end{figure*}

\vspace{-3mm}
\subsection{Datasets}
The experiments were selected from the public RSI semantic segmentation dataset ISPRS Potsdam\cite{isprs_potsdam}, competition dataset CVPR DGLC\cite{cvpr_dglc}, and Xiangtan dataset\cite{rs_ssl_seg_gl} from the Gaofen-2 satellite covering Xiangtan, China. The spatial resolution and number of ground object types of the three datasets are shown in TABLE \ref{tab1}.

\begin{table}[h]
\vspace{-5mm}
\renewcommand\arraystretch{1.5}
\caption{Datasets introduction\label{tab1}}
\centering
\setlength{\tabcolsep}{1mm}{\begin{tabular}{ccc}
\hline
Dataset Name& Spatial Resolution& Class Num\\
\hline
ISPRS Potsdam& 0.05m& 6\\
CVPR DGLC& 0.5m& 7\\
Xiangtan& 2m& 8\\
\hline
\end{tabular}}
\end{table}

\vspace{-3 mm}
\subsection{Experimental setup}
The experiments follow the general paradigm of SSL models \cite{simclr,moco,rs_ssl_cl} and are divided into two main steps: self-supervised pretraining and supervised fine-tuning. Self-supervised pretraining uses all unlabeled training set data, pretrained for 200 epochs with the batch size set to 256. Next, the entire model encoder is frozen, and a small amount of labeled data is used to fine-tune and train the decoder, with the labeled data selected as 1\% of the pretrained unlabeled data. The decoder is finally used to obtain the RSI semantic segmentation results.

To quantitatively explore the FSD of the FALSE, we conducted several experiments using the introduced confidence weights $\alpha$, with the positive sample pair similarity threshold $T$ set to 0.9. The results are presented in Experiment I.

On this basis, we selected SimCLR, which represents the original SSCL model; PCL, which represents the SSCL model that joins clusters; and the Barlow twins, which represent the SSCL model without constructing negative samples. These are used as the baseline tested on the ISPRS Potsdam, CVPR DGLC, and Xiangtan datasets, which are then compared with the FALSE model having a confidence weight of 1. The results are presented in Experiment II.

\begin{table}[h]
\vspace{-5mm}
\renewcommand\arraystretch{1.5}
\caption{Semantic segmentation IoU (\%) of 6 classes of ground objects in Potsdam dataset when 5 different $\alpha$ are selected\label{tab2}}
\centering
\setlength{\tabcolsep}{2mm}{\begin{tabular}{cccccc}
\hline
$\alpha$&	0& 0.3&	0.5& 0.7& 1\\
\hline
Imp. Surface& 47.87& 47.87& \underline{48.08}& 47.82& \textbf{48.16}\\	
Building& 47.15& \textbf{47.72}& 46.15& \underline{47.42}& 46.83\\	
Low Veg.& 41.22& 42.07& 42.06& \underline{42.22}& \textbf{42.38}\\	
Tree& 27.94& 27.37& \underline{30.31}& 28.05& \textbf{30.75}\\	
Car& \underline{28.22}& 24.86& 26.42& 27.05& \textbf{28.29}\\	
Clutter/background& 4.74& \underline{5.26}& 4.85& 4.93& \textbf{5.26}\\
\hline
\end{tabular}}
\end{table}

\begin{table}[h]
\vspace{-5mm}
\renewcommand\arraystretch{1.5}
\caption{Semantic segmentation IoU (\%) of 7 classes of ground objects in DGLC dataset when 5 different $\alpha$ are selected\label{tab3}}
\centering
\setlength{\tabcolsep}{2.5mm}{\begin{tabular}{cccccc}
\hline
$\alpha$&	0& 0.3&	0.5& 0.7& 1\\
\hline
Urban& 62.45& 61.69& \underline{63.39}& 62.3& \textbf{64.27}\\
Agriculture& 78.75& \underline{78.88}& 78.45& 78.61& \textbf{78.94}\\
Rangeland& \textbf{20.92}& 19.38& 18.91& 19.67& \underline{19.99}\\
Forest& 61.56& \textbf{63.86}& 61.36& 61.56& \underline{63.09}\\	
Water& 57.21& 58.15& 57.78& \underline{60.41}& \textbf{61.69}\\	
Barren& \textbf{46.43}& 43.56& 43.52& 44.43& \underline{45.39}\\	
Unknow& \textbf{96.12}& 95.54& 95.12& \underline{96.02}& 95.63\\
\hline
\end{tabular}}
\end{table}

\vspace{-3 mm}
\subsection{Experimental results and analysis}
\subsubsection{Experiment I, Analysis of confidence weight}
We selected five values that were uniformly distributed in the interval from 0 to 1: 0, 0.3, 0.5, 0.7, and 1. The experiments were repeated five times for each confidence weight while keeping the other parameters consistent.

Fig. \ref{fig_o} shows the semantic segmentation mIoU results of the FALSE model on the ISPRS Potsdam, CVPR DGLC, and Xiangtan datasets with five different confidence weights. The model performed best when fully trusting the possible FNS obtained by the model's FSD ($\alpha=1$). Compared with the original SSCL model, FALSE improves the mIoU by 0.72\% on the ISPRS Potsdam dataset, 0.8\% on the CVPR DGLC dataset, and 0.56\% on the Xiangtan dataset.
Moreover, when the model partially trusted the possible FNS obtained from the model's FSD ($0<\alpha<1$), the semantic segmentation performance became less stable. Considering that a confidence weight between 0 and 1 reduces the contribution of possible FNS to both positive and negative sample terms of the FNCC loss function, this phenomenon implies that the possible FNS (including FNS and MD-HNS) affects the stability of the semantic segmentation performance of the FALSE model.

\begin{table}
\vspace{-5mm}
\renewcommand\arraystretch{1.5}
\caption{Semantic segmentation IoU (\%) of 8 classes of ground objects in Xiangtan dataset when 5 different $\alpha$ are selected\label{tab4}}
\centering
\setlength{\tabcolsep}{2.5mm}{\begin{tabular}{cccccc}
\hline
$\alpha$&	0& 0.3&	0.5& 0.7& 1\\
\hline
Farmland& 63.82& \underline{64.10}& 63.82& 64.06& \textbf{64.45}\\
Urban& 0.00& 0.31& 0.53& \textbf{2.75}& \underline{0.96}\\
Rural areas& 16.62& 17.16& \underline{17.23}& \textbf{18.25}& 14.92\\
Water& 38.97& \underline{41.30}& 40.94& 40.42& \textbf{43.01}\\
Woodland& 78.27& \textbf{78.49}& 78.11& 78.12& \underline{78.36}\\
Grassland& \underline{1.95}& 1.86& 1.91& 1.70& \textbf{2.26}\\
Roads& 21.52& 21.30& 21.01& \textbf{21.68}& \underline{21.58}\\
Background& 97.56& 98.08& 98.09& \underline{98.2}& \textbf{98.23}\\
\hline
\end{tabular}}
\end{table}

TABLE \ref{tab2} - TABLE \ref{tab4} show the IoU of the FALSE model for various types of ground objects on the three datasets with five different confidence weights. Compared with five different confidence weight models, the FALSE model with confidence weight is 1 achieved the best mIoU for 5 of all 6 types of ground features on the Potsdam dataset, the best mIoU for 3 and the second-ranked mIoU for 3 of all 7 types of ground features on the DGLC dataset, the best mIoU for 4 and the second-ranked mIoU for 3 of all 8 types of ground features on the Xiangtan dataset.

\subsubsection{Experiment II, Comparison of 4 types of SSCL models for semantic segmentation}

\begin{table}[h]
\vspace{-5mm}
\renewcommand\arraystretch{1.5}
\caption{Semantic segmentation result of 4 different types of positive and negative sample construction strategies SSL models\label{tab5}}
\centering
\setlength{\tabcolsep}{0.7mm}{
\begin{tabular}{cccccccccc}
\hline
\multicolumn{1}{c}{\multirow{2}{*}{Method}} & \multicolumn{3}{c}{ISPRS Potsdam}                & \multicolumn{3}{c}{CVPR DGLC}                    & \multicolumn{3}{c}{Xiangtan}                    \\ \cline{2-10} 
\multicolumn{1}{c}{}                        & OA             & mIoU           & mAcc           & OA             & mIoU           & mAcc           & OA             & mIoU           & mAcc          \\ \hline
SimCLR                                      & 59.63          & 42.03          & \underline{54.26}          & \underline{81.10}          & \underline{60.49}          & \underline{70.41}          & \underline{79.20}          & 35.41          & 41.25         \\
PCL                                         & \underline{60.22}          & \underline{42.25}          & 54.20          & 77.36          & 54.09          & 62.56          & 77.85          & \underline{35.49}          & \underline{41.55}         \\
Barlow twins                                & 60.03          & 41.87          & 54.06          & 66.71          & 32.45          & 39.28          & 77.57          & 33.49          & 39.33         \\
FALSE(ours)                                 & \textbf{60.46} & \textbf{42.75} & \textbf{54.77} & \textbf{81.47} & \textbf{61.29} & \textbf{71.43} & \textbf{79.64} & \textbf{35.97} & \textbf{41.60} \\ \hline
\end{tabular}
}
\end{table}

TABLE \ref{tab5} shows the semantic segmentation results of the four different types of positive and negative sample construction strategy SSL models on the ISPRS Potsdam, CVPR DGLC, and Xiangtan datasets. The FALSE model achieves the best semantic segmentation results on these three datasets. Its Overall Accuracy (OA), mean Intersection-over-Union (mIoU), and mean class Accuracy (mAcc) outperform SimCLR, which represent the original SSCL model; PCL, which represents the SSCL model that joins clusters; and the Barlow twins, which represent the SSL model without constructing negative samples.

\vspace{-5mm}
\section{Conclusion and future work}
In this letter, we proposed the false negative sample aware contrastive learning model (FALSE) for the semantic segmentation of high-resolution RSIs. Under the restriction that self-supervised pretrained FNS are theoretically undecidable, the FALSE model achieves approximate determination of the FNS by coarse determination and precise calibration of FNS and quantitatively characterizes the ability of FNS self-determination (FSD) using confidence weights. Experiments on three RSI semantic segmentation datasets showed that FALSE effectively alleviates the SCE caused by SCI in the original SSCL of RSI. Compared with SimCLR, which represents the original SSCL model; PCL, which represents the SSCL model that joins clusters; and the Barlow twins, which represent the SSL model without constructing negative samples, FALSE improves mIoU by 0.7\% on average on ISPRS Potsdam, improves mIoU by 12.28\% on average on DGLC CVPR2018, and improves mIoU by 1.17\% on average on Xiangtan.

The current method is only a simple implementation of the model's FSD, introducing manually set confidence weights. Through the experiment, we found that the confidence weight corresponding to the best segmentation accuracy of different ground objects is not the same, so how to adjust the confidence weight adaptively for different ground objects in the dataset and give full play to the model's ability of FSD is a further issue to be considered in the future for the FALSE model.

\ifCLASSOPTIONcaptionsoff
  \newpage
\fi

\vspace{-5mm}
\bibliography{false_ref}

\begin{thebibliography}{10}
\providecommand{\url}[1]{#1}
\csname url@samestyle\endcsname
\providecommand{\newblock}{\relax}
\providecommand{\bibinfo}[2]{#2}
\providecommand{\BIBentrySTDinterwordspacing}{\spaceskip=0pt\relax}
\providecommand{\BIBentryALTinterwordstretchfactor}{4}
\providecommand{\BIBentryALTinterwordspacing}{\spaceskip=\fontdimen2\font plus
\BIBentryALTinterwordstretchfactor\fontdimen3\font minus
  \fontdimen4\font\relax}
\providecommand{\BIBforeignlanguage}[2]{{%
\expandafter\ifx\csname l@#1\endcsname\relax
\typeout{** WARNING: IEEEtran.bst: No hyphenation pattern has been}%
\typeout{** loaded for the language `#1'. Using the pattern for}%
\typeout{** the default language instead.}%
\else
\language=\csname l@#1\endcsname
\fi
#2}}
\providecommand{\BIBdecl}{\relax}
\BIBdecl

\bibitem{sup_class}
G.~Cheng, X.~Xie, J.~Han, L.~Guo, and G.-S. Xia, ``Remote sensing image scene
  classification meets deep learning: Challenges, methods, benchmarks, and
  opportunities,'' \emph{IEEE Journal of Selected Topics in Applied Earth
  Observations and Remote Sensing}, vol.~13, pp. 3735--3756, 2020.

\bibitem{sup_det}
K.~Li, G.~Wan, G.~Cheng, L.~Meng, and J.~Han, ``Object detection in optical
  remote sensing images: A survey and a new benchmark,'' \emph{ISPRS Journal of
  Photogrammetry and Remote Sensing}, vol. 159, pp. 296--307, 2020.

\bibitem{sup_seg0}
I.~Kotaridis and M.~Lazaridou, ``Remote sensing image segmentation advances: A
  meta-analysis,'' \emph{ISPRS Journal of Photogrammetry and Remote Sensing},
  vol. 173, pp. 309--322, 2021.

\bibitem{sup_seg}
H.~Li, K.~Qiu, L.~Chen, X.~Mei, L.~Hong, and C.~Tao, ``Scattnet: Semantic
  segmentation network with spatial and channel attention mechanism for
  high-resolution remote sensing images,'' \emph{IEEE Geoscience and Remote
  Sensing Letters}, vol.~18, no.~5, pp. 905--909, 2020.

\bibitem{m_contrastive}
X.~Zhao, R.~Vemulapalli, P.~A. Mansfield, B.~Gong, B.~Green, L.~Shapira, and
  Y.~Wu, ``Contrastive learning for label efficient semantic segmentation,''
  \emph{Proceedings of the IEEE/CVF International Conference on Computer
  Vision}, pp. 10\,623--10\,633, 2021.

\bibitem{sup_cha1}
Q.~Yuan, H.~Shen, T.~Li, Z.~Li, S.~Li, Y.~Jiang, H.~Xu, W.~Tan, Q.~Yang,
  J.~Wang \emph{et~al.}, ``Deep learning in environmental remote sensing:
  Achievements and challenges,'' \emph{Remote Sensing of Environment}, vol.
  241, p. 111716, 2020.

\bibitem{sup_cha2}
A.~Vali, S.~Comai, and M.~Matteucci, ``Deep learning for land use and land
  cover classification based on hyperspectral and multispectral earth
  observation data: A review,'' \emph{Remote Sensing}, vol.~12, no.~15, p.
  2495, 2020.

\bibitem{m2_contrastive}
S.~Saha, M.~Shahzad, L.~Mou, Q.~Song, and X.~X. Zhu, ``Unsupervised
  single-scene semantic segmentation for earth observation,'' \emph{IEEE
  Transactions on Geoscience and Remote Sensing}, 2022.

\bibitem{rs_ssl_cl}
C.~Tao, J.~Qi, W.~Lu, H.~Wang, and H.~Li, ``Remote sensing image scene
  classification with self-supervised paradigm under limited labeled samples,''
  \emph{IEEE Geoscience and Remote Sensing Letters}, 2020.

\bibitem{rs_ssl_seg_gl}
H.~Li, Y.~Li, G.~Zhang, R.~Liu, H.~Huang, Q.~Zhu, and C.~Tao, ``Global and
  local contrastive self-supervised learning for semantic segmentation of hr
  remote sensing images,'' \emph{IEEE Transactions on Geoscience and Remote
  Sensing}, 2022.

\bibitem{simclr}
T.~Chen, S.~Kornblith, M.~Norouzi, and G.~Hinton, ``A simple framework for
  contrastive learning of visual representations,'' \emph{International
  conference on machine learning}, pp. 1597--1607, 2020.

\bibitem{data_aug}
O.~Henaff, ``Data-efficient image recognition with contrastive predictive
  coding,'' \emph{International Conference on Machine Learning}, pp.
  4182--4192, 2020.

\bibitem{moco}
K.~He, H.~Fan, Y.~Wu, S.~Xie, and R.~B. Girshick, ``Momentum contrast for
  unsupervised visual representation learning,'' \emph{2020 IEEE/CVF Conference
  on Computer Vision and Pattern Recognition (CVPR)}, pp. 9726--9735, 2020.

\bibitem{contralea_with_simclr}
Y.~Tian, D.~Krishnan, and P.~Isola, ``Contrastive multiview coding,''
  \emph{European conference on computer vision}, pp. 776--794, 2020.

\bibitem{contralea_with_simclr2}
P.~Bachman, R.~D. Hjelm, and W.~Buchwalter, ``Learning representations by
  maximizing mutual information across views,'' \emph{Advances in neural
  information processing systems}, vol.~32, 2019.

\bibitem{pcl}
J.~Li, P.~Zhou, C.~Xiong, and S.~C. Hoi, ``Prototypical contrastive learning of
  unsupervised representations,'' 2021.

\bibitem{deepcluster}
M.~Caron, P.~Bojanowski, A.~Joulin, and M.~Douze, ``Deep clustering for
  unsupervised learning of visual features,'' \emph{Proceedings of the European
  conference on computer vision (ECCV)}, pp. 132--149, 2018.

\bibitem{swav}
M.~Caron, I.~Misra, J.~Mairal, P.~Goyal, P.~Bojanowski, and A.~Joulin,
  ``Unsupervised learning of visual features by contrasting cluster
  assignments,'' \emph{Advances in Neural Information Processing Systems},
  vol.~33, pp. 9912--9924, 2020.

\bibitem{imbalanced}
W.~Xia, C.~Ma, J.~Liu, S.~Liu, F.~Chen, Z.~Yang, and J.~Duan, ``High-resolution
  remote sensing imagery classification of imbalanced data using multistage
  sampling method and deep neural networks,'' \emph{Remote Sensing}, vol.~11,
  no.~21, p. 2523, 2019.

\bibitem{barlowtwins}
J.~Zbontar, L.~Jing, I.~Misra, Y.~LeCun, and S.~Deny, ``Barlow twins:
  Self-supervised learning via redundancy reduction,'' \emph{International
  Conference on Machine Learning}, pp. 12\,310--12\,320, 2021.

\bibitem{byol}
J.-B. Grill, F.~Strub, F.~Altch{\'e}, C.~Tallec, P.~Richemond, E.~Buchatskaya,
  C.~Doersch, B.~Avila~Pires, Z.~Guo, M.~Gheshlaghi~Azar \emph{et~al.},
  ``Bootstrap your own latent-a new approach to self-supervised learning,''
  \emph{Advances in Neural Information Processing Systems}, vol.~33, pp.
  21\,271--21\,284, 2020.

\bibitem{no_negative_method}
X.~Chen and K.~He, ``Exploring simple siamese representation learning,''
  \emph{IEEE Conference on Computer Vision and Pattern Recognition}, 2021.

\bibitem{infoNCE}
A.~Van~den Oord, Y.~Li, and O.~Vinyals, ``Representation learning with
  contrastive predictive coding,'' \emph{arXiv e-prints}, pp. arXiv--1807,
  2018.

\bibitem{isprs_potsdam}
F.~Rottensteiner, G.~Sohn, J.~Jung, M.~Gerke, C.~Baillard, S.~Benitez, and
  U.~Breitkopf, ``The isprs benchmark on urban object classification and 3d
  building reconstruction,'' \emph{ISPRS Annals of the Photogrammetry, Remote
  Sensing and Spatial Information Sciences I-3 (2012), Nr. 1}, vol.~1, no.~1,
  pp. 293--298, 2012.

\bibitem{cvpr_dglc}
I.~Demir, K.~Koperski, D.~Lindenbaum, G.~Pang, J.~Huang, S.~Basu, F.~Hughes,
  D.~Tuia, and R.~Raskar, ``Deepglobe 2018: A challenge to parse the earth
  through satellite images,'' \emph{IEEE Conference on Computer Vision and
  Pattern Recognition Workshops}, 2018.

\end{thebibliography}

\end{document}